\documentclass[conference]{IEEEtran}
\usepackage{graphicx}
\usepackage{amsmath}
\usepackage{microtype}
\usepackage{hyperref}
\usepackage{cleveref}

\begin{document}
\title{Spiking Analog VLSI Neuron Assemblies \\ as Constraint Satisfaction Problem Solvers}

\author{\IEEEauthorblockN{Jonathan Binas, Giacomo Indiveri, Michael Pfeiffer}
\IEEEauthorblockA{Institute of Neuroinformatics, University of Zurich and ETH Zurich, Switzerland\\
Email: jbinas@ini.ethz.ch}
}

\maketitle

\begin{abstract}
 Solving constraint satisfaction problems (CSPs) is a notoriously expensive computational task.
 Recently, it has been proposed that efficient stochastic solvers can be obtained through appropriately configured spiking neural networks performing Markov Chain Monte Carlo (MCMC) sampling.
 The possibility to run such models on massively parallel, low-power neuromorphic hardware holds great promise; however, previously proposed networks are based on probabilistically spiking neurons, and thus rely on random number generators or external noise sources to achieve the necessary stochasticity, leading to significant overhead in the implementation.
 Here we show how stochasticity can be achieved by implementing deterministic models of integrate and fire neurons using subthreshold analog circuits that are affected by thermal noise.
 We present an efficient implementation of spike-based CSP solvers using a reconfigurable neural network VLSI device, and the device's intrinsic noise as a source of randomness.
 To illustrate the overall concept, we implement a generic Sudoku solver based on our approach and demonstrate its operation.
  We establish a link between the neuron parameters and the system dynamics, allowing for a simple temperature control mechanism.

\end{abstract}

 \section{Introduction}

 Constraint satisfaction problems (CSPs) include some of the most prominent problems in science and engineering.
 A CSP is defined by a set of variables and a set of conditions (constraints) on those variables that need to be satisfied simultaneously.
 Solutions to a given CSP typically form a vanishingly small subset of an exponentially large search space, rendering this type of problem NP-hard in the general case.
 Many CSP solvers therefore involve problem-specific heuristics to avoid searching the entire space.
 Alternatively, one can consider a massively parallel system of simple computing elements that tackle different parts of the problem simultaneously, and discover a global solution through communicating local search mechanisms.
 Here, we explore event-based neural hardware as a substrate of computation, however, there are various other approaches, such as quantum annealing \cite{Farhi2001}, special cellular automata hardware \cite{Fujita2010}, or oscillator networks \cite{Mostafa2015}.
 The idea of using recurrent neural networks for finding solutions to computationally hard problems goes back to Hinton \cite{Hinton1984} and Hopfield \cite{Hopfield1985}.
 While the deterministic Hopfield network fails in the general case because of local minima it can get caught in, the probabilistic Boltzmann machine proposed by Hinton and Sejnowski \cite{Hinton1984} overcomes this limitation by sampling its states from a probability distribution rather than evolving along a deterministic trajectory.

 It has been shown recently, how such stochastic samplers can be implemented in networks of spiking neurons \cite{Buesing2011,Maass2014}.
 The resulting \emph{neural sampling} framework has been applied to constraint satisfaction problems by Jonke et al. \cite{Jonke2014}, demonstrating
 advantages of the spiking network approach over standard Gibbs sampling in certain cases.
 The main technological advantage of using such neural samplers in practical applications lies in the ability to implement spiking neurons efficiently in neuromorphic hardware.
 However, the models proposed in \cite{Buesing2011,Jonke2014} are difficult to directly transfer onto spiking VLSI neurons, because they require individual neurons to emit spikes probabilistically, sampling from a probability distribution of a specific form. This is not easily implementable in an electronic circuit without explicit sources of noise, or random number generators.
 Here we show, for the first time, how a similar sampling mechanism can be implemented in a system of analog electronic integrate and fire neurons.
 To avoid the additional cost of implementing dedicated noise sources on chip, we propose a mechanism that makes use of the small amount of (thermal) noise that is present in any analog electronic system to achieve the desired stochastic network dynamics.
 We demonstrate these principles using a standalone system, based on a reconfigurable neuromorphic processor \cite{Qiao2015}, comprising a configurable network of adaptive exponential integrate and fire neurons and synapse circuits with biophysically realistic dynamics \cite{Chicca2014}.
 Once programmed for a given CSP, the system will output streams of candidate solutions.
 Our experimental results demonstrate that these samples predominantly represent configurations with no or few violated constraints.

 \section{Solving CSPs with Spiking Neural Networks}

 Solving an NP-complete decision problem stochastically through sampling means transforming it into an NP-hard optimization problem, and solving it using some kind of annealing mechanism rather than sophisticated algorithmics.
 Thereby, a cost function is formulated such that the solutions to a given problem are transformed into optima.
 It is not clear whether the corresponding optimization problem is easier to solve in general, however, typically the conversion can be done in polynomial time, and in some cases the optimization problem can be parallelized more easily or more efficiently than the decision problem.
 Many types of problems can be transformed to simple graph structures  with little effort \cite{Lucas2014}, and thus almost naturally map onto a network of nodes that interact through positive and negative links.
 In contrast to most conventional algorithmic solvers, the optimization-based approach does not depend on problem-specific heuristics, and can thus be regarded as more general.

 As a first step, we outline how arbitrary discrete CSPs can be mapped onto a network of neurons.
 We consider constraint satisfaction problems defined by a set of $n$ discrete variables $\{x_1,\ldots,x_n\}$ on finite domains and a set of constraints, each linking several of those variables, e.g. $(x_i=a \lor x_j=b) \land x_i \neq x_j$ etc.
 Without loss of generality, any such problem can be expressed in terms of binary variables by using a one-hot scheme, i.e. by representing each discrete variable $x_i$ as a vector of binary variables $(x_{i,k})_k$, where exactly one is active at any time, $x_i=a \Leftrightarrow x_{i,a}=1,~x_{i,k \neq a}=0$.
 Furthermore, a CSP can be written in conjunctive normal form, i.e. in the form $\land_i \left(\lor_j l_{ij}\right)$, where the $l_{ij}$ are literals (binary variables or their negations).

 In a network of spiking neurons, following the models introduced in \cite{Buesing2011,Nessler2013}, the state of each variable is represented by the spiking activity of multiple cells.
 There is one cell per value a variable can assume, and a cell is called active at time $t$ if it has emitted a spike within a certain time window $[t-\tau_\mathrm{bin},t]$.
 The neural sampling framework \cite{Buesing2011} describes how a network of stochastically spiking cells can generate samples from a Boltzmann distribution, where the samples are represented by network states.
 A problem in conjunctive normal form can be mapped to an Ising model \cite{Choi2010} and therefore be solved by the spiking network sampling from the corresponding Boltzmann distribution.
 Alternatively, special network motifs implementing the OR relation between several variables can be used to achieve a smaller and more efficient network \cite{Jonke2014}.
 The energy function which defines the probability distribution of outputs, i.e. the sampled states of all variables, is designed such that solutions to the CSP occur at particularly high rates.
 Note that such a sampler does not know whether its current state represents a solution and will continue exploring the state space.
 However, potential solutions can be validated by a secondary mechanism in polynomial time, for example by testing their conformity with the pairwise interactions between cells in the network.

\begin{figure}[t]
  \centering
  \includegraphics[scale=.8]{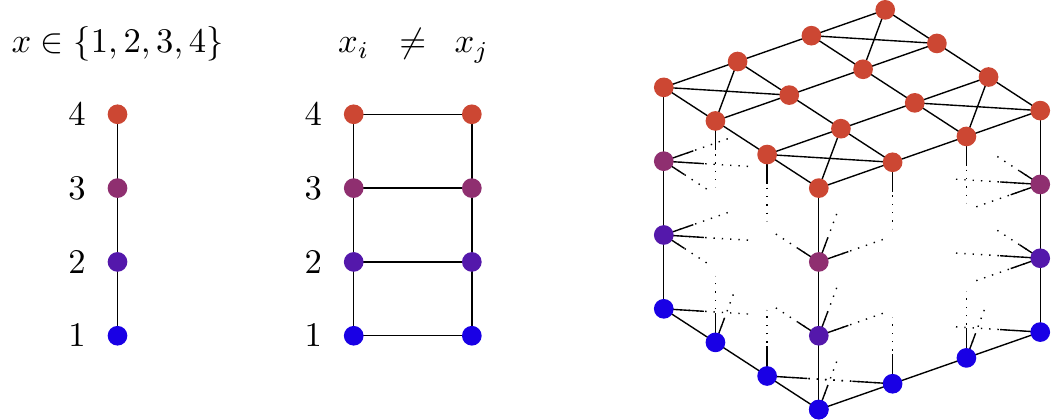}
  \caption{
 Illustration of the Sudoku solver network. Colored dots represent neurons, lines crossing several dots represent all-to-all inhibitory connections between them.
 A variable \(x\in\{1,2,3,4\}\) is one-hot encoded and its value represented by activity of one of four cells, which mutually inhibit each other (left). The condition \(x_i \neq x_j\) can be implemented by setting negative connections between cells representing the same value (middle).
 The whole Sudoku network can be implemented using constraints of the type \(x_i \neq x_j\) (right; only a subset of the 64 cells comprising the $4\times4\times4$ cube is shown).
 A $9\times9$ Sudoku solver implemented based on the same scheme would require $9^3=729$ neurons.
  }
  \label{fig:network}
\end{figure}

 To illustrate our experimental results we implemented a $4\times4$ Sudoku solver on spiking analog hardware. A reduced version of the standard $9\times9$ Sudoku problem was chosen due to the limited number of neurons available on the device.
 In this example problem, there are 16 variables $x_i \in \{1,2,3,4\},\,i=1,\ldots,16$ that are aligned in a $4\times4$ grid and restricted by the constraints that no two variables in a row, in a column, or in a $2\times2$ quadrant must assume the same value.
 As described above, this can be written in binary form, by introducing four binary variables for each variable $x_i$, whereby exactly one of them must be true at any time.
 Whenever one cell becomes active it shuts off all the others representing the same variable for a certain period by providing a strong inhibitory post-synaptic potential (IPSP) of duration $\tau_\mathrm{inh}$.
 On the other hand, all cells in the array receive a constant excitatory input current, such that one of them will spike if none are active, ensuring that the respective variable is in a defined state at any time.
 The activity of a single cell is limited by the refractory period, inactivating a cell for a duration $\tau_\mathrm{ref}$ immediately after a spike is emitted.
 The implementation of the Sudoku solver is illustrated in \cref{fig:network}.
 The constraint $x_i \neq x_j$, which exists for any two variables of the same row, column, or $2\times2$ quadrant, is implemented by specifying inhibitory interactions between cells representing the same value (\cref{fig:network} middle), such that the two variables cannot assume the same value at the same time.

 \section{Stochastic Dynamics in an Analog VLSI Neural Network}

 In this section we introduce a simple neuron model that can be used to describe our analog VLSI implementation of neural sampling.
 In principle, the stochastic spiking neurons used in previous work \cite{Buesing2011,Nessler2013,Jonke2014} can be approximated by integrate-and-fire neurons, which are injected large amounts of noise, as proposed by \cite{Merolla2010,Neftci2013,Petrovici2013}.
 This approach, however, requires an independent noise source for every cell, and therefore cannot easily be implemented directly in hardware.
 Instead, we propose a mechanism that is based on conventional deterministic neuron models, and becomes stochastic through slight jitter in the 
 duration of temporally extended pulses in analog VLSI.
 Such small (thermal) fluctuations are inherent to any analog electronic system, and thus can be exploited in analog hardware implementations of the proposed model.
 
\begin{figure}[h]
  \centering
  \includegraphics[scale=.8, trim=0 0 0 -.4cm]{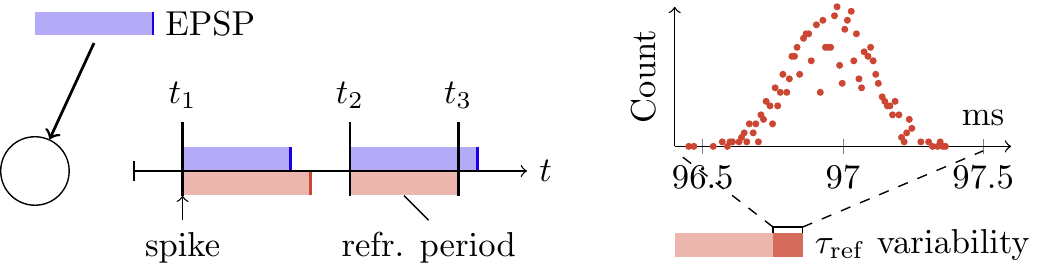}
  \caption{
 Abstract neuron model.
 EPSPs (blue) are provided to a cell at times $t_1$ and $t_2$, triggering spikes and subsequent state transition of the cell into refractory period (red).
 Due to the variability in the duration of the refractory period and post-synaptic potentials, sometimes one or the other is longer, even though on average, they are roughly of the same length.
 For instance, the EPSP provided at $t_2$ lasts slightly longer than the refractory period triggered by the spike at $t_2$, leading to a second spike at $t_3$.
 Equivalent effects are observed for IPSPs (functionally, the refractory period is equivalent to an IPSP in our model).
 On the right, measurements of refractory period duration from the actual hardware are shown, indicating an approximately Gaussian distribution.
  }
  \label{fig:model1}
\end{figure}

 To illustrate our model, we assume simple leaky integrate-and-fire neurons that produce an output spike and remain in refractory period for a duration $\tau_\mathrm{ref}$ when their membrane potential crosses a threshold $\Theta$.
 A spike in one cell triggers an excitatory or inhibitory postsynaptic potential (PSP) at synapses connecting it to other cells.
 In the simplest case, which is also at the core of previous models \cite{Buesing2011,Nessler2013}, this can be thought of as a rectangular signal of duration $\tau_\mathrm{inh}$ or $\tau_\mathrm{exc}$, respectively.
 We make the assumption that the magnitudes of these signals are large enough to either trigger a spike in the target cell almost instantaneously (for excitatory inputs), 
 or silence the cell completely (for inhibitory inputs), such that additional excitatory inputs have no effect.
 Note that the refractory period can be thought of as a strong inhibitory input of a cell to itself.
 The stochasticity in our system is then introduced by small amounts of noise in the duration of those PSPs and refractory periods.
 As a consequence, we can regard $\tau_\mathrm{inh}$, $\tau_\mathrm{exc}$, and $\tau_\mathrm{ref}$ as mean values, and in practice the durations are jittered around those values, as shown in \cref{fig:model1}.

 As an example, assume two neurons that are coupled through inhibitory connections and are driven by a constant external current, and assume further that $\tau_\mathrm{ref}\ll\tau_\mathrm{inh}$.
 In this circuit, whichever cell became active first would keep inhibiting the other cell. This is due to the short refractory period, which lets the active neuron spike again before the IPSP it provides to the other cell ends.
 This network would end up in a local optimum, and would never explore the other possible state where the second cell is active.
 If, however, the refractory period $\tau_\mathrm{ref}$ and the inhibitory pulse width $\tau_\mathrm{inh}$ are of similar size, small amounts of noise in the analog system, leading to jitter in the duration of both pulses, will sometimes cause the inhibitory PSP to be longer than the refractory period, and vice versa. Such a system could indeed explore all possible states.
 This mechanism can be considered a kind of noise amplification or, alternatively the system can be regarded as being close to a critical point, where vanishingly small fluctuations can lead to dramatically different behavior.
 Intuitively, longer refractory periods cause more explorative behavior, whereas short refractory periods let the network settle into local energy minima.
 Note that all time constants in the system are defined relative to each other, and the relation to real time is of little relevance.
 Thus, extending the refractory period is equivalent to shortening the PSPs, i.e. weakening the links between nodes.
 In that sense, the refractory period can be regarded as a kind of temperature, that could be used in an annealing schedule to steer the dynamics of the network.

 The assumptions made to construct our model are fulfilled in intrinsically noisy analog neural hardware which can, with this method, be configured such that very small fluctuations in the electronic signals can lead to large deviations in the network dynamics. 
 The jitter in the refractory period or PSP durations is thereby introduced by thermal fluctuations in the analog signals representing those variables.
 As signals in analog electronics are affected by fabrication-induced variability, in practice systematic deviations will be observed between PSPs of different synapses.
 In order for the sampling mechanism to function properly under these conditions, the variability in PSP duration due to thermal noise needs to be at least of the same order as the variability due to fabrication-induced mismatch. 
 If this is not the case, the distribution gets biased and might not represent the problem accurately enough. Note that this is not necessarily problematic as long as the minima are conserved.
 For our experimental setup, we used the programmable neuromorphic device described in \cite{Qiao2015}, comprising 256 integrate and fire neurons and 128k programmable synapses, to implement a network that evolves in real time and produces a stream of output events that can be interpreted as states of the system.
 The network described in \cref{fig:network} was programmed into the hardware by setting the respective inhibitory connections and run by injecting small amounts of direct current into each cell.
 The mean values of the tunable time constants $\tau_\mathrm{ref}$ and $\tau_\mathrm{inh}$ were set to approximately 100~ms for all cells.
 The sampling rate at which the network states were evaluated was set to 10~Hz, such that the network activity was binned over 100~ms for each sample.
 While the system could be run much faster, the time constants were set to relatively large values to allow for visualization of the network evolution in real time.

 \section{Experimental Results}

\begin{figure}[tb]
  \centering
  \includegraphics[scale=.8]{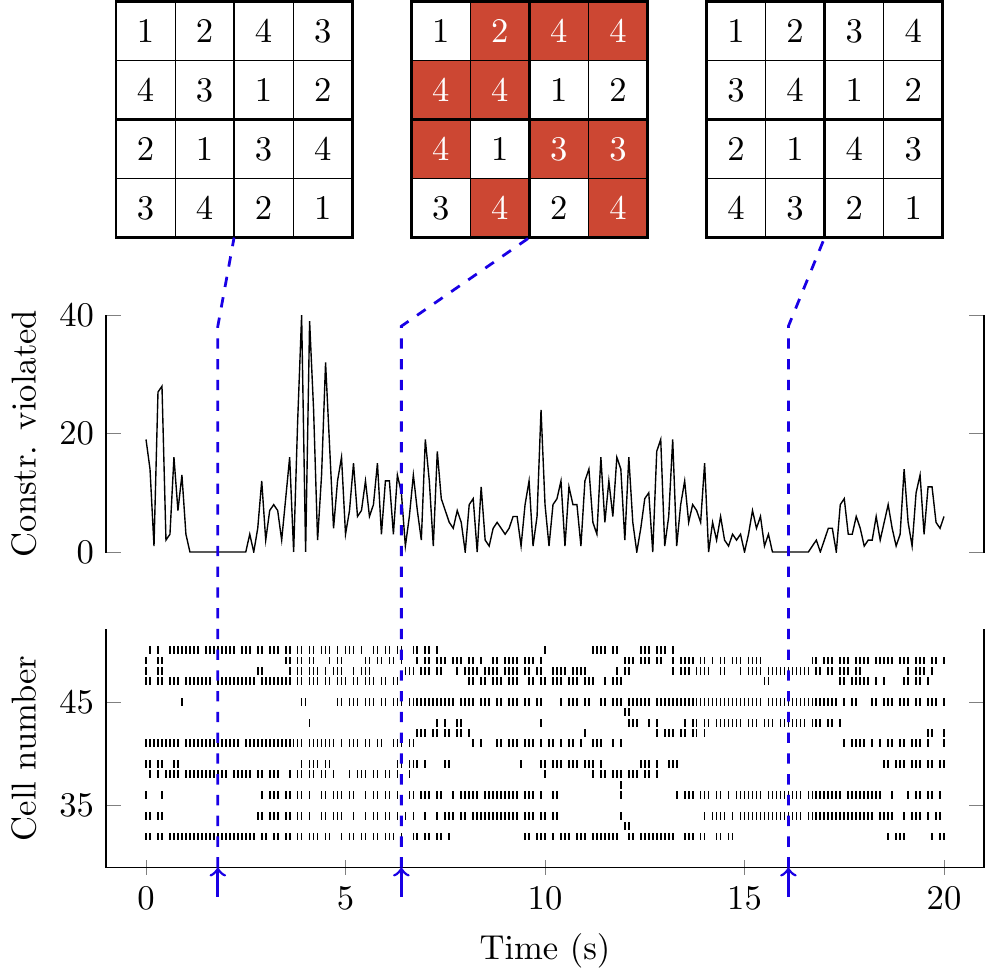}
  \caption{
    Sudoku solver on an analog VLSI chip.
 The panels show representative spiking activity of a subset of the neurons used in the Sudoku solver implementation (bottom), and the temporal evolution of the number of constraints violated (top) over a period of 20~s.
 A binary sample vector is acquired by binning the network activity over 100~ms and assigning a 1 or 0 to each cell, depending on whether it has spiked in the given interval or not.
 Three example states are shown, where a red cell indicates that the respective variable violates one or more constraints.
 The system frequently converges to states that represent solutions to the problem (0 constraints violated), but, due to the noise it is able to escape from those energy minima.
 Note that the solutions found at around 2~s and 16~s are not identical (3 and 4 are swapped).
  }
  \label{fig:results1}
\end{figure}

 \Cref{fig:results1} shows representative spiking activity of the Sudoku solver network implemented and running on an analog VLSI chip. The system occasionally converges to states solving the problem (0 constraints violated), however, it is also able to escape from those local optima and explore other possible states.
 The ``temperature'' or ``exploration rate'', can be controlled by tuning the neuron parameters, i.e. the IPSP duration or the length of the refractory period.
 As expected from our considerations above, the system can be constrained to lower energy regions by lowering the temperature parameter, i.e. decreasing the refractory period.
 \begin{figure}[t]
  \centering
  \includegraphics[scale=.8]{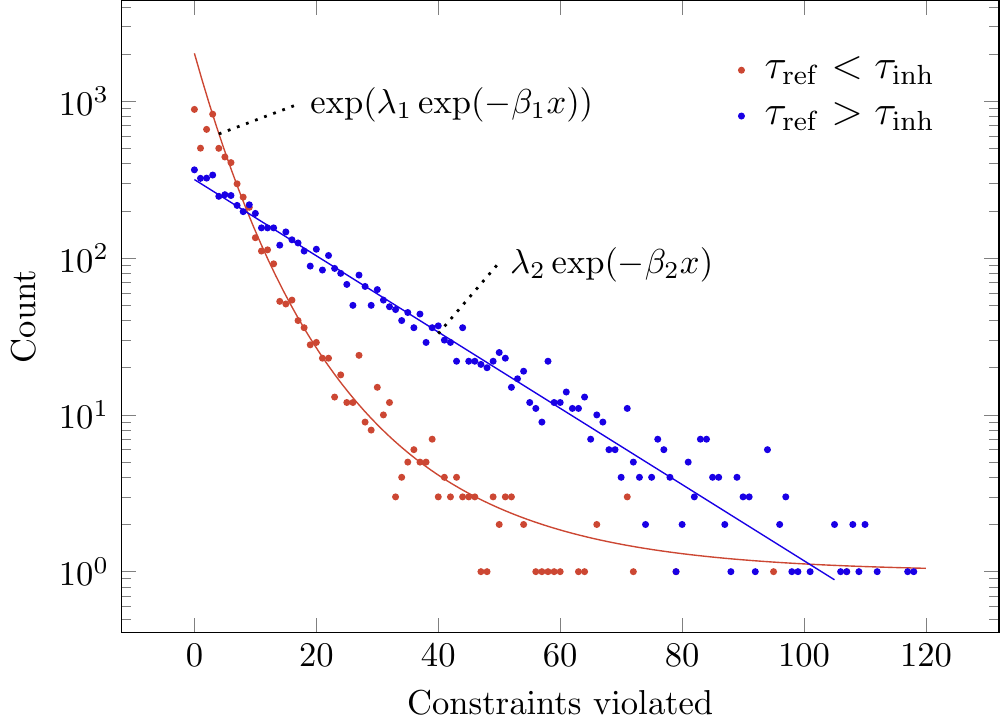}
  \caption{
 Empirical analysis of the samples generated by the hardware CSP solver.
 The plots show histograms of the number of constraints violated, based on 10~mins of spiking activity of the Sudoku network.
 Solid lines are least-squares fits to the data.
 Slightly varying one of the network parameters (the mean refractory period, in this case) leads to fundamentally different behavior of the system.
 If the refractory period is longer than the IPSP, the frequency of states of a certain energy decreases exponentially with the number of constraints violated.
 For shorter refractory periods, the system samples almost exclusively from very low energy states, and the distribution can be described by a double exponential function.
 For these measurements, $\tau_\mathrm{inh}$ was fixed to $\approx110$~ms, while $\tau_\mathrm{ref}$ was set to $\approx100$~ms (red) or $\approx120$~ms (blue).
  }
  \label{fig:resultsHist1}
\end{figure}
  \Cref{fig:resultsHist1} shows the distribution of the number of constraints violated, i.e. the measure of ``energy'' that we intend to minimize.
 As expected, we observe fundamentally different behavior for the case where the refractory period is larger than the IPSP, compared to the case where it is smaller. This leads to a phase transition-like phenomenon when this threshold is crossed. 
 For longer refractory periods the distribution can be well fitted by an exponential function, indicating a strong concentration around the low energy states.
 If the refractory period is shorter than the IPSP, however, the distribution is even more concentrated around zero, and can be approximated by a double exponential function.
 As shown in \cref{fig:resultsHist2}, a similar effect can be observed in the distribution of energy jumps, i.e. the difference in the number of constraints violated between one state and the next. Good fits are again obtained by exponential and a double exponential functions, respectively.
 This indicates that, similar to a Gibbs sampler, the system predominantly switches between states of similar energy, rather than doing high energy jumps.

 \section{Conclusion}

 We present the first analog VLSI implementation of a CSP solver based on neural sampling with spiking neurons.
 Our contribution is a simple neuron model that achieves stochasticity without the external noise sources required by previous approaches, but instead exploits small variations and noise in the duration of temporally extended signals.
 The empirical results obtained from an analog neuromorphic processor demonstrate the function and performance of the proposed mechanism.
 While the hardware system used in our experiments is deliberately slowed down to operate at timescales similar to real neurons, our approach could, without restrictions, be used with much faster hardware to solve computationally hard problems quickly and efficiently.
 We can empirically relate the duration of the refractory period, or alternatively the duration of PSPs, to a temperature parameter, such that those time constants could be varied in an annealing schedule to control the temperature.
 We regard our work as a proof-of-concept and further research is required to optimize performance and analyze theoretical properties of the model.

\begin{figure}[t]
  \centering
  \includegraphics[scale=.8]{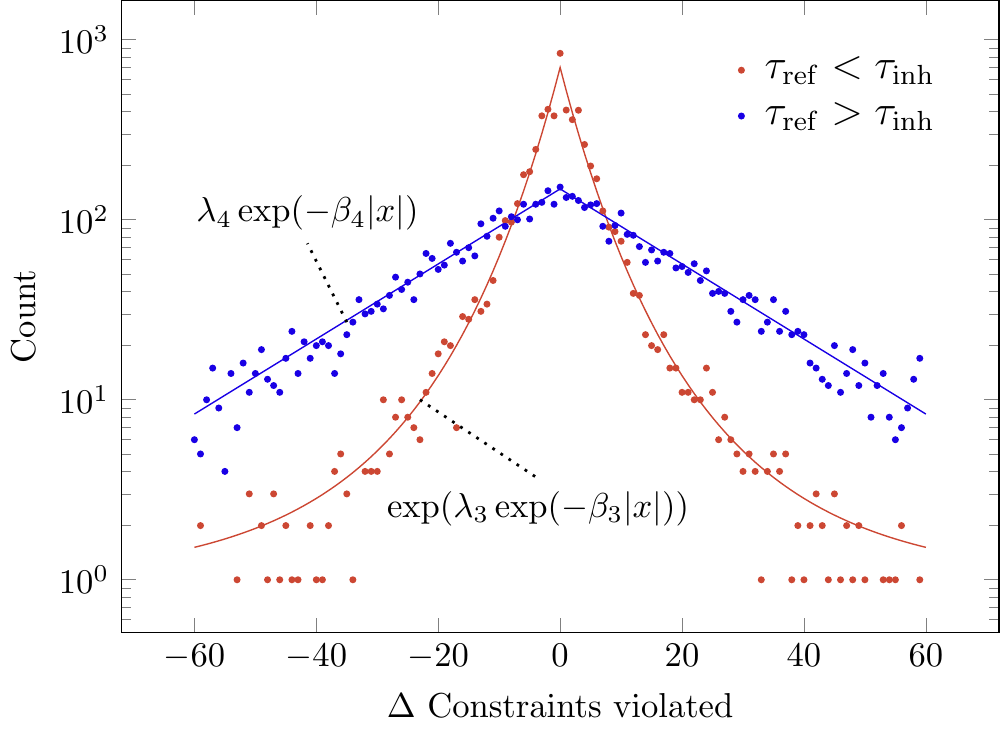}
  \caption{
 Empirical analysis of the dynamics of the analog VLSI Sudoku solver.
 The plots show the distribution of energy jumps, i.e. the difference in the number of constraints violated between two consecutive states for different network parameters (values as in \cref{fig:resultsHist1}).
  }
  \label{fig:resultsHist2}
\end{figure}

\section*{Acknowledgment}

 We thank Wolfgang Maass and our colleagues at the Institute of Neuroinformatics for fruitful discussions.
 The research was supported by the Swiss National Science Foundation Grant 200021\_146608 and by the EU ERC Grant “neuroP” (257219).

\bibliography{bibliography}
\bibliographystyle{IEEEtran}

\end{document}